\documentclass[sigconf]{acmart}
\usepackage{algorithm}
\usepackage{algorithmic}
\usepackage{xcolor,colortbl}
\usepackage{multirow, booktabs}

\usepackage{setspace}
\usepackage{amsmath}
\usepackage{amsthm}
\usepackage{amsfonts, mathtools}
\usepackage{bm}
\usepackage{comment}
\usepackage{graphicx}
\usepackage{caption}
\usepackage{subcaption}
\usepackage[colorinlistoftodos]{todonotes}

\newcommand{\hmu}{\hat{\mu}}

\usepackage{newfloat}
\usepackage{listings}
\floatstyle{ruled}
\newfloat{listing}{tb}{lst}{}
\floatname{listing}{Listing}

\newcommand{\R}{\mathbb{R}}

\DeclareMathOperator*{\argmax}{arg\,max}

\newtheorem{proposition}{Proposition}
\newtheorem{lemma}{Lemma}

\newtheorem{assumption}{Assumption}

\newtheorem{definition}{Definition}

\AtBeginDocument{%
  \providecommand\BibTeX{{%
    \normalfont B\kern-0.5em{\scshape i\kern-0.25em b}\kern-0.8em\TeX}}}

\copyrightyear{2022}
\acmYear{2022}
\acmConference[WWW '22] {Proceedings of the ACM Web Conference 2022}{April 25--29, 2022}{Virtual Event, Lyon, France.}
\acmBooktitle{Proceedings of the ACM Web Conference 2022 (WWW '22), April 25--29, 2022, Virtual Event, Lyon, France}

\acmPrice{15.00}
\acmDOI{10.1145/3485447.3512103}
\acmISBN{978-1-4503-9096-5/22/04} 

\begin{document}

\title{LBCF: A Large-Scale Budget-Constrained Causal Forest Algorithm}


\author{Meng Ai$^{1}$, Biao Li$^{1}$, Heyang Gong$^{1,2}$, Qingwei Yu$^{1}$, Shengjie Xue$^{1}$, Yuan Zhang$^{1}$, Yunzhou Zhang$^{1}$, Peng Jiang$^{1}$}
\affiliation{%
 \institution{$^1$Kuaishou Inc., China}
 \institution{$^2$University of Science and Technology of China, China}
 \country{}
}
\affiliation{%
  \institution{\{aimeng, libiao, gongheyang03, yuqingwei,
  xueshengjie, zhangyuan13, zhangyunzhou, jiangpeng\}@kuaishou.com}
}

\renewcommand{\shortauthors}{Meng Ai et al.}

%

\begin{abstract}



Offering incentives (e.g., coupons at Amazon, discounts at Uber and video bonuses at Tiktok) to user is a common strategy used by online platforms to increase user engagement and platform revenue. 
Despite its proven effectiveness, these marketing incentives incur an inevitable cost and might result in a low ROI (Return on Investment) if not used properly. On the other hand, different users respond differently to these incentives, for instance, some users never buy certain products without coupons, while others do anyway. 
Thus, how to select the right amount of incentives (i.e. treatment) to each user under budget constraints is an important research problem with great practical implications. 
In this paper, we call such problem as a \textit{budget-constrained treatment selection} (BTS) problem.

The challenge is how to efficiently solve BTS problem on a Large-Scale dataset and achieve improved results over the existing techniques. We propose a novel tree-based treatment selection technique under budget constraints, called  \textit{Large-Scale Budget-Constrained Causal Forest} (LBCF) algorithm, which is also an efficient treatment selection algorithm suitable for modern \textit{distributed computing} systems. A novel offline evaluation method is also proposed to overcome an intrinsic challenge in assessing solutions' performance for BTS problem in randomized control trials (RCT) data. We deploy our approach in a real-world scenario on a large-scale video platform, where the platform gives away bonuses in order to increase users' campaign engagement duration.
The simulation analysis, offline and online experiments all show that our method outperforms various tree-based state-of-the-art baselines \footnote{Our code is available on  \href{https://github.com/www2022paper/WWW-2022-PAPER-SUPPLEMENTARY-MATERIALS}{github:} \emph{https://github.com/www2022paper/WWW-2022-PAPER-SUPPLEMENTARY-MATERIALS}}.
The proposed approach is currently serving over hundreds of millions of users on the platform and achieves one of the most tremendous improvements over these months.

\end{abstract}

\ccsdesc[300]{Information systems~Personalization}
\ccsdesc[500]{Mathematics of computing~Linear programming}



\keywords{Personalization, Heterogeneous causal effects, Constraint optimization,
Treatment Selection, Distributed Computing}

\maketitle


\section{Introduction}
\label{intro}

Conducting marketing campaigns by giving away incentives is a popular and effective way \cite{reutterer2006dynamic} 
to boost user engagement and platform revenue, for example, discounts \cite{lin2017monetary}  in ride-sharing (Uber \cite{ zhao2019uplift}, Didi) and
coupons in e-commerce (Alibaba \cite{zhao2019unified}, Booking.com \cite{goldenberg2020free, makhijani2019lore}). 

In industrial settings, these marketing campaigns are under limited budget constraints, thus it is crucial to distribute limited incentives with efficiency. And the most challenging task is to identify target user with heterogeneity, 
i.e., different users respond differently to various levels of coupons (e.g., 10\% off, 20\% off, etc.). In other words, some users never buy certain products without these coupons, while others do anyway.
Thus, it is an important research problem to investigate how we should allocate a limited incentive budget to users with heterogeneity in order to maximize the overall return (e.g. user engagement, platform revenue, etc.). Recent studies begin to solve this problem with causal analysis frameworks, where incentives and return are regarded as ``treatments'' and ``outcomes'', respectively. In this work, we regard such problem as a \textit{budget-constrained treatment selection} (BTS) problem.



\textbf{Methodology}. BTS problem can be solved by many existing techniques. The challenge is how to efficiently solve BTS problem on a large-scale dataset and achieve improved results over the existing techniques. In this paper, we mainly focus on the \textit{tree-based} techniques because of its excellent performance in industry. These techniques can be classified into two categories.

The first category is \textit{uplift random forest with greedy treatment selection}. After obtaining the estimation of heterogeneous treatment effects (treatment effect and uplift are exchangeable in this paper) for all treatments, these solutions, e.g., uplift random forests on Contextual Treatment Selection (CTS) ~\cite{zhao2017uplift}, simply select the treatment with the maximal treatment effect for each user \cite{guelman2015uplift, radcliffe2011real} . We call such treatment selection policy as a greedy treatment selection policy. In Section \ref{4.1.1}, by a toy example, we show that such greedy treatment selection policy is sub-optimal for BTS problem.

\begin{figure*}[]
    \centering
    \begin{subfigure}[b]{0.48\textwidth}
      \includegraphics[width=\textwidth]{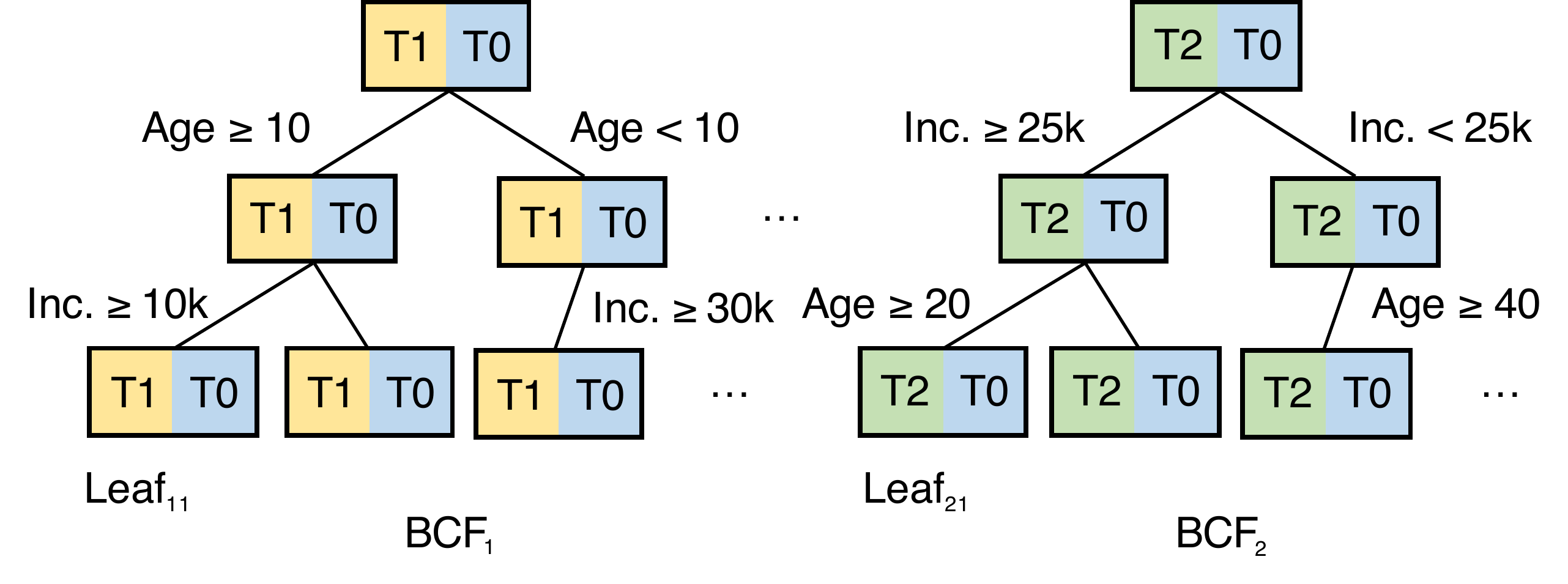}
      \caption{}
      \label{fig:causalforest}
    \end{subfigure}%
    ~
    \begin{subfigure}[b]{0.48\textwidth}
      \includegraphics[width=\textwidth]{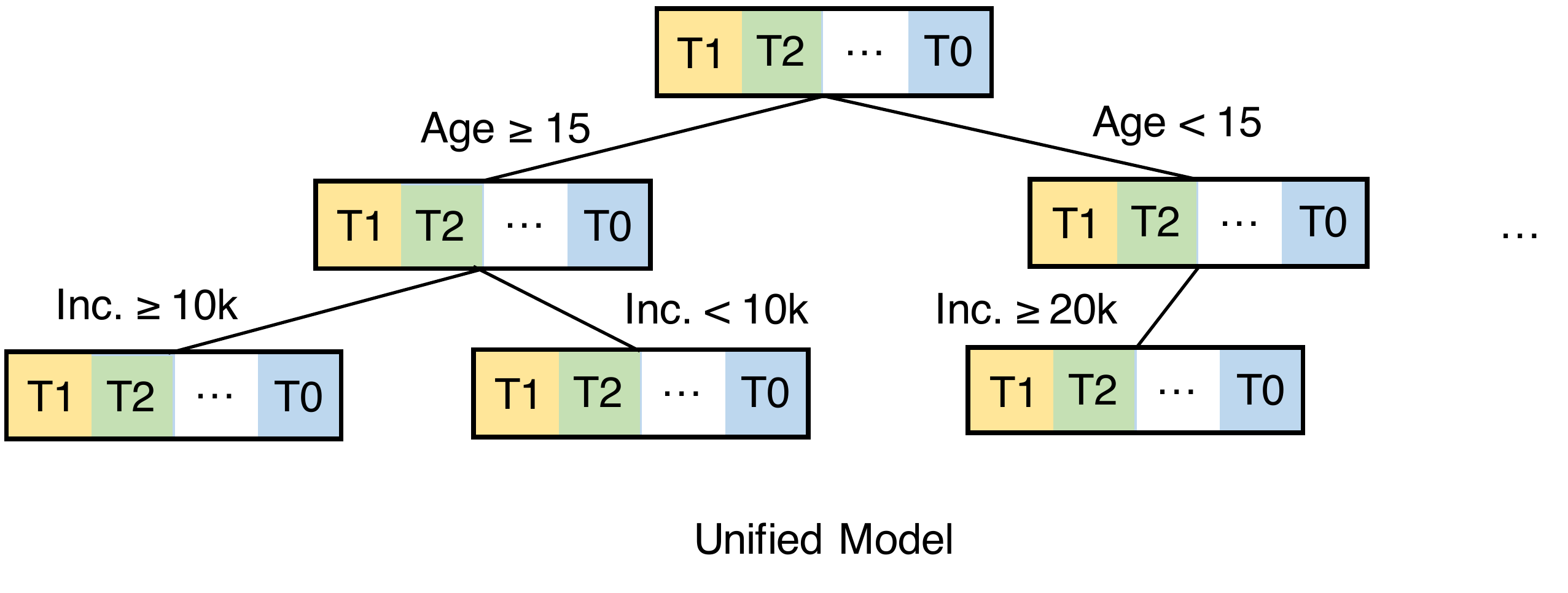}
      \caption{}
      \label{fig:multiple}
    \end{subfigure}
    \caption{$T_0$ stands for control group, and $T_{i}$ stands for i-th treatment group. (a) A MBCF model consists of several binary models. (b) The proposed UDCF model with multiple treatments. }
    \label{fig:binarycomparemultiple}
    \vspace{-0.1cm}
\end{figure*}

The second category is the \textit{optimal treatment selection algorithm} recently proposed by \citet{tu2021personalized}. This algorithm has two limitations especially for a large-scale BTS problem. First, on a large-scale dataset, the author suggested the cohort-level optimization instead of the member-level optimization because of the lack of a large-scale linear programming optimization solver. Second, in order to realize multi-treatment effect estimation with Binary Causal Forest (see \citet{athey2018generalized}) (BCF), \citet{tu2021personalized} proposed to train multiple binary causal forest (MBCF) separately where each BCF corresponds to one treatment. Nonetheless, in this divide-and-conquer approach, a user might belong to different leaf nodes in each binary model (see Fig. \ref{fig:causalforest}). Consequently, we are looking at different feature space when estimating the treatment effect or uplift with regard to different treatments (see Fig.  \ref{fig:roi_a}). Thus, the obtained treatment effects are not rigorously comparable across treatments. Besides, it is computationally consuming and hard to maintain if we have multiple models serving at the same time. 

In this paper, we propose a Large-Scale Budget-Constrained Causal Forest (LBCF) algorithm to overcome the above limitations. LBCF consists of two components: member-level treatment effect estimation and budget-constrained optimization. For the first component, we design a novel splitting criteria that allows similar users from multiple treatment groups to reside in the same node. 
The proposed splitting criteria allows us to deal with multiple treatments with a unified model (Fig.  \ref{fig:multiple}), so that the treatment effect w.r.t different incentive levels can be estimated within the same feature space (Fig.  \ref{fig:roi_b}). To this end, in addition to inter-node heterogeneity, our proposed approach also encourages the intra-node treatment effect heterogeneity to facilitate the downstream budget-constrained optimization task. For the other component, after obtaining users' treatment effect to different incentive levels, we can then formulate the budget-constrained optimization task as a multi-choice knapsack problem (MCKP)~\cite{sinha1979multiple, kellerer2004multidimensional, martello1990knapsack}.
Although MCKP has been studied for decades (e.g., the classic linear-time approximation algorithm, Dyer-Zemel algorithm \cite{dyer1984n}), 
existing optimization methods are not designed for modern infrastructure, especially distributed computing frameworks, such as Tensorflow \cite{abadi2016tensorflow} 
and Spark. As a result, these methods are not scalable to today's large-scale online platforms. In this work, we leverage the convexity of the Lagrange dual of MCKP and design an efficient parallelizable bisection search algorithm, called Dual Gradient Bisection (DGB). Compared with other state-of-the-art approximation optimization methods, DGB can be deployed in distributed computing systems much more easily with the same $O(N)$ time complexity and requires no additional hyperparameter tuning as its gradient descent-based alternative.


\textbf{Policy Evaluation}. Offline evaluation for BTS is another challenging problem. Existing evaluation methods for BTS problem suffer from several limitations. One type of methods, such as AUUC \cite{gutierrez2017causal, rzepakowski2012decision, soltys2015ensemble}, Qini-curve \cite{radcliffe2007using} and AUCC \cite{du2019improve}, in multi-treatment case, rank all users according to the score, which is the maximal predicted treatment effect among all possible treatment assignments. However, the to-be-evaluate policy does not necessarily select the maximal treatment. Another method is the expected outcome metric proposed by Zhao et al.~\cite{zhao2017uplift}. In this metric, the evaluated users are not the whole RCT users, which causes the consumed budget to change with different treatment selection policies. To overcome the limitations of these existing evaluation methods, we propose a novel evaluation metric, called the Percentage Mean Gain (PMG) for BTS problem. Our metric enables a more holistic assessment of treatment selection policies. 


\textbf{Dataset and Test}. In order to fully verify the performance of our LBCF algorithm, we conduct three kinds of tests: simulation test, offline test on a real-word dataset and online AB test. 

In simulation test, We use the same method as \citet{tu2021personalized} to generate the synthetic RCT dataset. In order to make a convincing result, we also use the same measurement metric defined in \citet{tu2021personalized}: normalized mean of individualized treatment effect (ITE). we compare our proposed LBCF algorithm with five tree-based baseline methods: uplift random forests \cite{guelman2015uplift, radcliffe2011real} on Euclidean Distance (ED), Chi-Square (Chi) and Contextual Treatment Selection (CTS), all of which are from CausalML package; Causal tree with stochastic optimization (CT.ST) and Causal forest with deterministic optimization (CF.DT), both of which are the top two methods recommended by \citet{tu2021personalized}. The simulation results show the good performance of LBCF algorithm under different levels of noise.

In offline test, we first collected a web-scale RCT dataset from an video streaming platform. The dataset records the users' campaign engagement duration (i.e., ``outcome'') in seven randomly enrolled incentive groups, each offering bonuses at different levels (i.e., ``multi-treatment''). With our released complete dataset and proposed evaluation protocol PMG, more researchers in our community can be potentially involved in related studies.

We further deploy the proposed approach on the platform. Online A/B experiments show that in terms of campaign engagement duration, our algorithm can significantly outperform the baseline methods by at least $0.9\%$ under the same budget, which is a tremendous improvement over these months. The scalability of our LBCF algorithm is also demonstrated in online experiments. Currently, the LBCF algorithm is serving over hundreds of millions of users.

To sum up, the contribution of this paper can be summarized as follows:



\begin{figure}[]
    \centering
    \begin{subfigure}[b]{0.24\textwidth}
      \includegraphics[width=\textwidth]{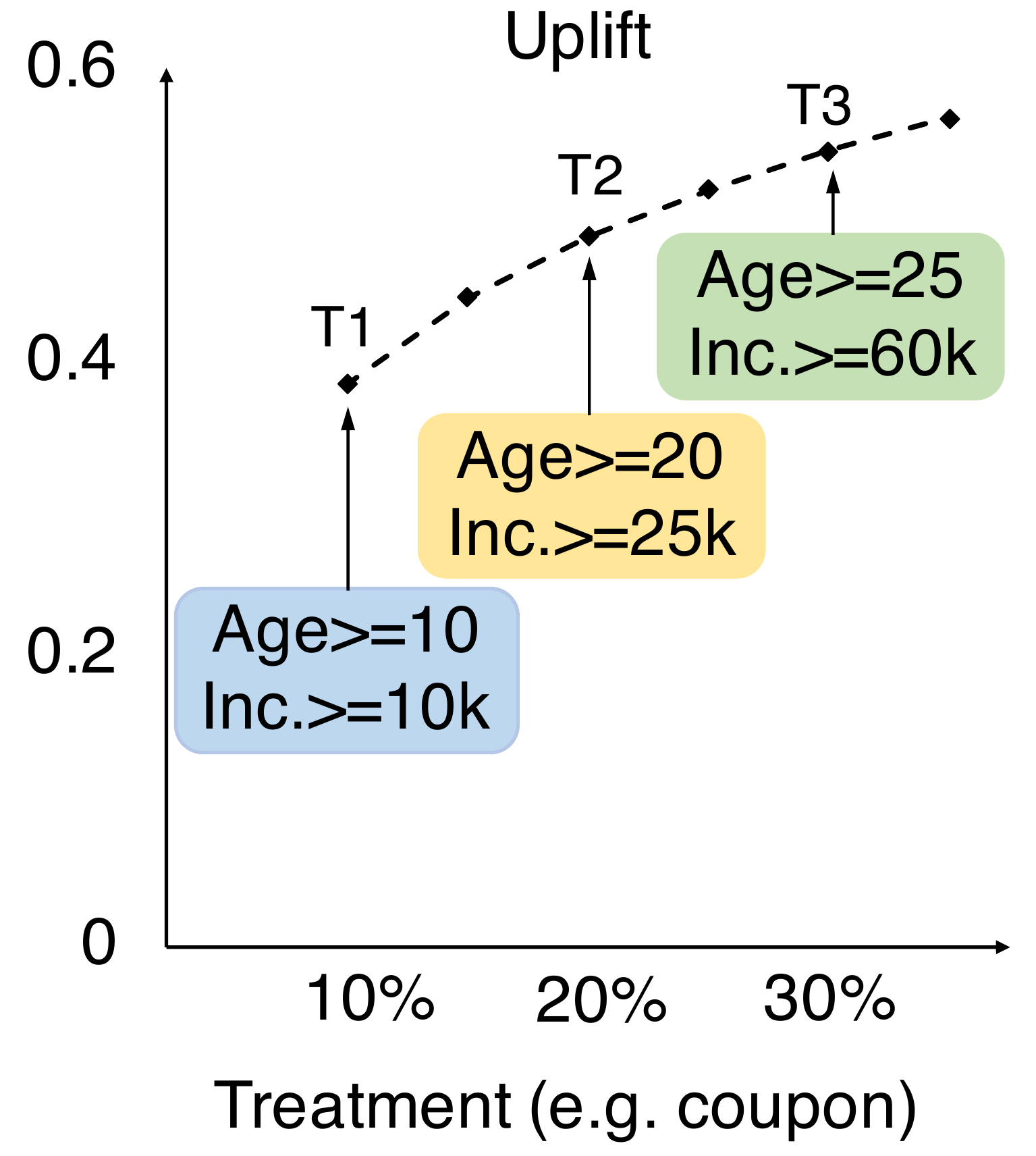}
      \caption{}
      \label{fig:roi_a}
    \end{subfigure}%
    ~
    \begin{subfigure}[b]{0.24\textwidth}
      \includegraphics[width=\textwidth]{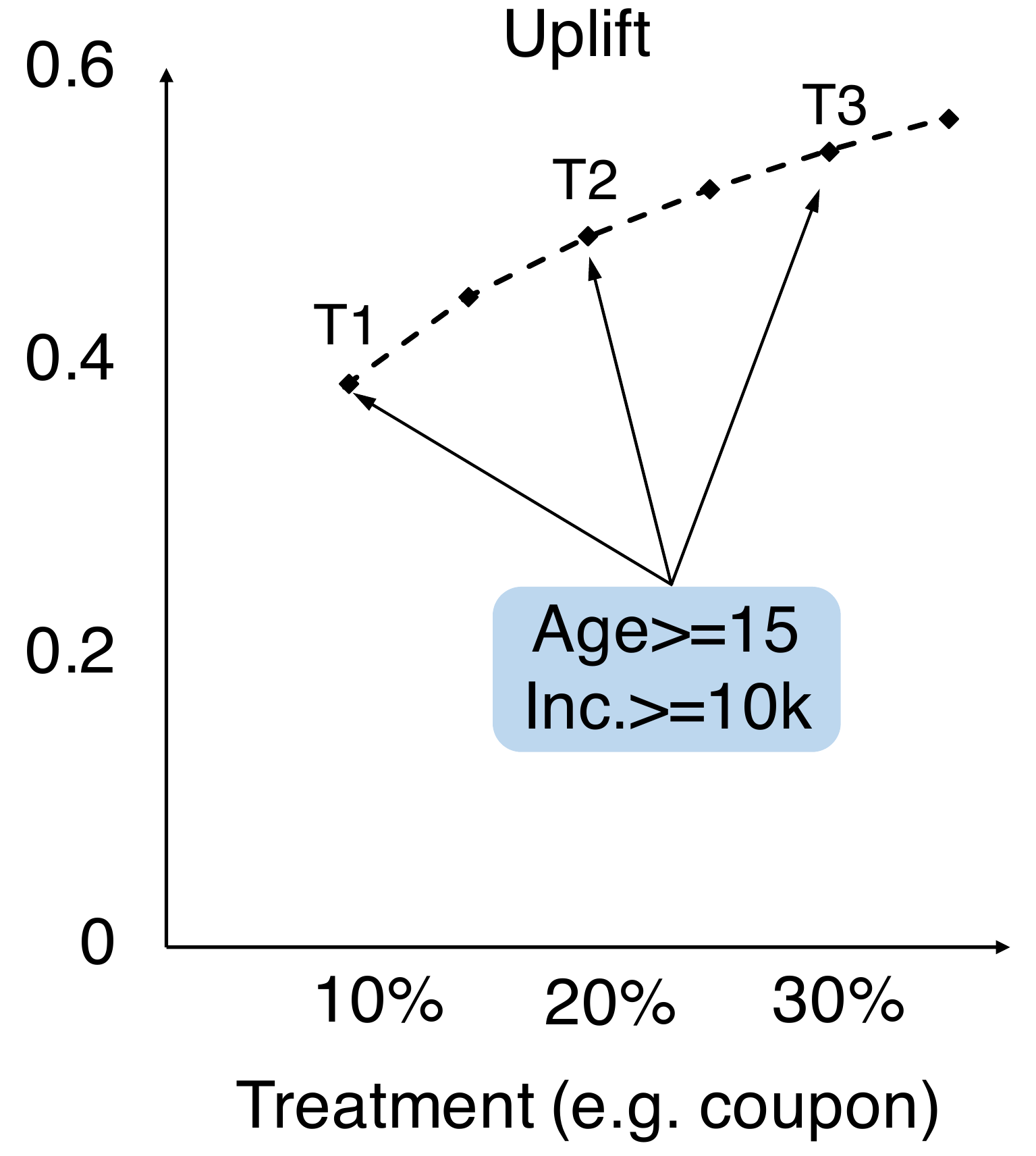}
      \caption{}
      \label{fig:roi_b}
    \end{subfigure}
    \caption{(a) Uplifts with three treatments $T_1$, $T_2$ and $T_3$ estimated by three models are not comparable because they belong to three feature space. (b) Uplifts with three treatments $T_1$, $T_2$ and $T_3$ estimated by one model are comparable because they belong to the same feature space.}
    \label{fig:roi}
    \vspace{-0.35cm}
\end{figure}




\begin{itemize}
    \item We propose a novel tree-based treatment selection technique under budget constraint, called Large-Scale Budget-Constrained Causal Forest (LBCF) algorithm. The proposed approach has been deployed on a real-world large-scale platform serving over hundreds of millions of users.
    \item We propose a novel offline evaluation method for budget-constrained treatment selection (BTS) problem, called the percentage mean gain (PMG), which tackles the intrinsic limitations in assessing BTS problem's solutions with offline datasets collected with randomized control trials (RCT).
    \item We conduct a series of extensive experiments, which includes a simulation test on a public synthetic dataset, an offline test on collected a real-world dataset and an online AB test on a large-scale video platform. Results demonstrate the effectiveness of our proposed approach and also proves its scalability to large-scale industrial application scenarios.
\end{itemize}

\section{Problem Formulation}
\label{prelim}

In this work, we focus on maximizing the overall return while deciding how to offer the incentives in order to comply with the global budget constraint. 

We adopt the potential outcomes framework \cite{rubin1974estimating, splawa1990application} 
to express the treatment effects of the incentives on return, where incentives and return are regarded as ``treatments'' and ``outcomes'', respectively. We use upper case letters to denote random variables and lower case letters their realizations. We use boldface for vectors and normal typeface for scalars. 


\begin{itemize}
	\item $\mathbf{X}$ represents the feature vector and $\mathbf{x}$ its realization. Let $\mathbb{X}^d$ denote the $d$-dimensional feature space. 
	
	\item $T$ represents the treatment. We assume there are $K$ mutually
exclusive treatments encoded as $\{1,\ldots,K \}$. The control group is indicated by $T=0$. 
	
	\item Let $Y$ denote the response and $y$ its realization. Throughout this paper we assume the larger the value of $Y$, the more desirable the outcome. 
\end{itemize}

Suppose we have a data set of size $N$ containing the joint realization of $(\mathbf{X}, T, Y)$ collected from RCT. We use superscript $(i)$ to index the samples, $ \big(\, \mathbf{x_i}, t_i, y_i  \,\big), i=1,\ldots,N.$ $t_i \in \{0, 1, ..., K\}$.
We posit the existence of potential outcomes $Y_i(T_i=j)$ and $Y_i(T_i=0)$
corresponding to the outcome we would have observed given the treatment assignment $T_i=j$ or $T_i=0$ respectively, and try to estimate the conditional average treatment effect (CATE) function assume that they have the same feature values $\mathbf{x_i}$:
\begin{equation}
\label{cate}
 \theta_{ij}(\mathbf{x_i})=\mathbb{E}[Y_i(T_i=j)-Y_i(T_i=0)|\mathbf{X_i}=\mathbf{x_i}].
\end{equation}

 The unconfoundedness assumption (i.e., the treatment assignment is as-if random once we control for the features $\mathbf{X_i}$) and SUTVA assumption \cite{rosenbaum1983central} are assumed as usual for estimating $\theta_{ij}$.  
In addition, we consider the random variables $C$, representing the cost associated with the treatment. We assume that there are no cost if $T=0$ and the cost $c_{ij}$ for applying each treatment $j$ to each user $i$ is known beforehand. Let $B$ denote the total budget. Then given $(\mathbf{X}, T, Y, C)$ and manually-set budget $B$, our goal is to maximize the total return by learning an optimal treatment selection policy $\pi_B$, which includes all the $z_{ij}$. Note $z_{ij} \in \{1, 0\}$ is the treatment assignment variable if $\pi_B$ selects treatment $j$ to user $i$ or not and boldface $\mathbf{z}$ is the vectored version.

\section{Policy Evaluation}
\label{3.1}
The first challenge for BTS problem is how to evaluate the solution, which is a difficult task due to missing counterfactual outcome, e.g., we cannot observe the potential outcome change (i.e. treatment effect) for a customer if we select treatment $A$ (e.g. 10\% off coupon) instead of treatment $B$ (e.g. no coupon). In this section, we propose a novel evaluation method for BTS problem.


\textbf{Limitation of Existing Methods.}
Two kinds of evaluation methods have been studied recently. 

The first one is to use the metric \textit{area under the uplift curve} (AUUC)~\cite{rzepakowski2012decision} to evaluate treatment effect with user data gathered from randomized control trials (RCT)~\cite{kohavi2013online, kohavi2014seven, tang2010overlapping, xu2015infrastructure}. In multi-treatment case, such methods require all users to be ranked in descending order by their score, which is the maximal predicted treatment effect among all possible treatment assignments. However, the to-be-evaluate policy does not necessarily select the maximal treatment because the maximal treatment may not be the optimal one.

\begin{figure}[]
    \centering
    \resizebox{\columnwidth}{!}{%
    \begin{subfigure}[b]{0.28\textwidth}
      \includegraphics[width=\textwidth]{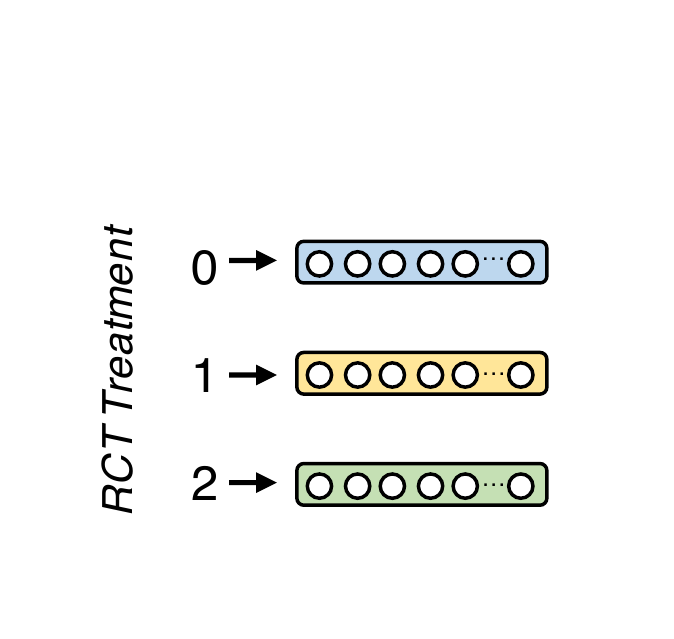}
      \caption{}
      \label{fig:overlap_samples_a}
    \end{subfigure}%
    \hspace{0.03\textwidth}
    ~
    \begin{subfigure}[b]{0.28\textwidth}
      \includegraphics[width=1\textwidth]{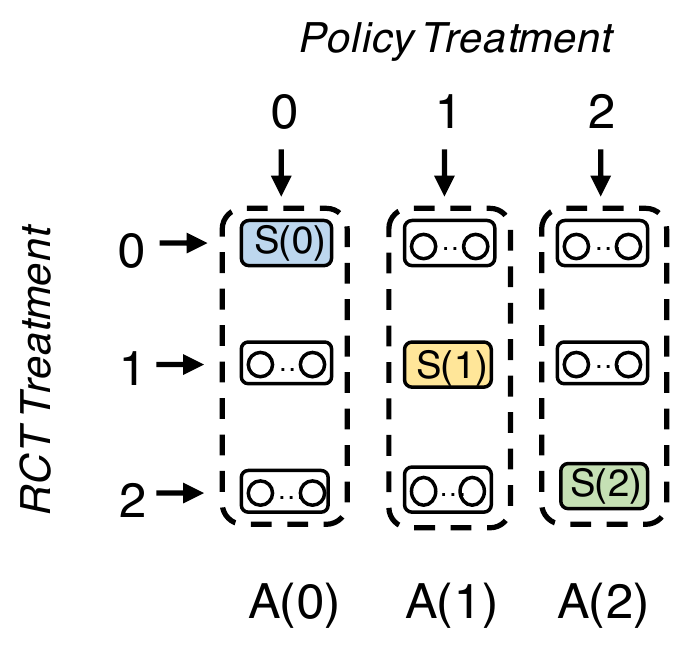}
      \caption{}
      \label{fig:overlap_samples_b}
    \end{subfigure}}
    \caption{Definition of Terms: (a) RCT Treatment (b) Policy Treatment, Overlapping Subset $S(j)$ and Policy Subset $A(j)$.}
    \label{fig:overlap_samples}
\end{figure}

The second one is the Expected Outcome metric propose by \cite{zhao2017uplift}. It estimates the expected outcome by summing the weighted outcome of users where the RCT's and policy's treatments (Fig. \ref{fig:overlap_samples}) match. For our BTS problem, the main limitations of this metric is that the evaluated users is not the whole RCT users. Therefore the Expected Outcome is not the mean outcome of the whole RCT users, which causes the consumed budget change with different treatment selection policies. Two policies with different consumed budgets are not comparable.

Considering the limitations of the two methods above, we propose a novel policy evaluation metric, called Percentage Mean Gain (PMG) for BTS problem.

\textbf{Definitions.} Suppose we have a sample set of size $N$ containing the joint realization of $(\mathbf{X}, T, Y)$ collected from RCT. A hollow circle represents a sample in Fig. \ref{fig:overlap_samples}. Firstly we give some definitions:
\begin{definition}[RCT Treatment]
\label{rt}
The \textit{RCT treatment} is the original treatment randomly assigned by RCT. See Fig. \ref{fig:overlap_samples_a}.
\end{definition}

\begin{definition}[Policy Treatment]
\label{st1}
The \textit{Policy Treatment} is the final treatment selected by the policy $\pi_B$. See Fig. \ref{fig:overlap_samples_b}.
\end{definition}

\begin{definition}[Overlapping Subset]
Overlapping Subset, denoted as $S(j)$, is a set containing samples whose \textit{RCT treatment} and \textit{policy treatment} both are $j$. For example, $S(0)$, $S(1)$ and $S(2)$ in Fig. \ref{fig:overlap_samples_b}.
\end{definition}

Therefore, for \textit{RCT treatment} every sample's outcome can be observed, but for \textit{policy treatment} only the outcome of those samples which are in the \textit{overlapping subset} can be observed.

\begin{definition}[Policy Subset]
\label{ss}
Policy subset, denoted as $A(j)$, is a set containing samples whose \textit{policy treatment} are $j$, such as $A(0)$, $A(1)$ and $A(2)$ in Fig. \ref{fig:overlap_samples_b}.
\end{definition}

\textbf{Proposed Method.}

To evaluate a policy $\pi_B$ under given budget $B$, we need to calculate the expected outcome of all RCT samples (which is different from the Expected Outcome metric in \cite{zhao2017uplift}) if they were under \textit{policy treatment}, denoted as $E(Y|T=\pi_B)$. Obviously, $$E(Y|T=\pi_B) = \frac{\sum_{j=0}^K |A(j)|E(Y|T=j,\mathbf{X} \in A(j))}{N}.$$ 
However, for \textit{policy treatment}, we can only observe the actual outcome of those samples in \textit{overlapping subset}. Thanks to the randomness of RCT, hereby each \textit{overlapping subset} $S(j)$ can be regarded as a randomly selected subset from each \textit{policy subset} $A(j)$. 
Thus
\begin{proposition}
\label{theo:unbias}
The sample average $\bar{y}(j) = \sum_{S(j)} y_i / |S(j)|$ is an unbiased estimate for $E(Y|T=j,\mathbf{X} \in A(j))$.
\end{proposition}

Therefore, $E(Y|T=\pi_B)$ can be estimated by $\frac{1}{N}\sum_{j=0}^K |A(j)|\bar{y}(j)$. To measures the outcome's percentage mean gain (PMG) by the policy $\pi_B$, we define a new metric:
\begin{equation}
\label{mapg}
 PMG = \frac{\frac{1}{N}\sum_{j=0}^K |A(j)|\bar{y}(j) - \hmu_{0} }{\hmu_{0}}.
\end{equation}
where $\hmu_{0}$ is the average outcome of samples in the control group. Obviously, the PMG is a plug-in estimator of the quantity $E[Y(T=\pi_B) - Y(T=0)]/E[Y(T=0)]$.

\section{Methodology}
\label{method}

We perform a comparison of several common tree-based techniques to solve BTS problem and introduce a novel approach called large-scale budget-constrained causal forest (LBCF) algorithm.

\subsection{Limitation of Existing Methods.}
\label{4.1}

\subsubsection{Uplift Random Forest.}
\label{4.1.1}

For treatment selection, most uplift random forest methods simply choose the treatment with the maximal treatment effect. We call such treatment selection policy as a greedy treatment selection policy. In the following, by a toy example, we show that such greedy treatment selection policy is sub-optimal under a given budget $B$.

We take ROI greedy treatment selection policy as an example. Define $ROI_{ij} = \theta_{ij}/({c_{ij} - c_{i0}})$, where $c_{i0}$ is assumed to be $0$ for simplicity. Under a giver budget $B$, ROI greedy policy is as follows:

\begin{itemize}
  \item [(i)] For each user $i$, select the treatment with max $ROI_{i{j^*}}$.
  \item [(ii)] Sort all users according to $ROI_{i{j^*}}$ in a descending order. 
  \item [(iii)] Pick user from top to bottom until $B$ is consumed up.
\end{itemize}


However, such ROI greedy policy is not optimal from the perspective of budget constraint. As in Table \ref{fig:allocation1}, suppose we have six users and the total budget $B=6$. Then the total treatment effect value for ROI greedy policy is $92$, that is to select treatment $1$ for $user_1$, treatment $2$ for $user_2$, treatment $2$ for $user_{3}$, and treatment $1$ for $user_{4}$. However, we can find another policy to get a larger value if we select treatment $2$ for $user_1$, select no treatment for $user_{4}$ and select the same for others as before. Then the budget is still $B=6$, but the total value increases to $98$.

\begin{table}[t]
    \centering
     \caption{ROI greedy policy: total treatment effect value = $92$. Another better policy is to select treatment $2$ for $user_1$ and select nothing for $user_{4}$, then the total value = $98$.}
     \label{fig:allocation1}
    \resizebox{\columnwidth}{!}{%
\begin{tabular}{@{}cccccccc@{}}
\toprule

                        & \multicolumn{3}{c}{Treatment 1}                                                     & \multicolumn{3}{c}{Treatment 2}                                                     & \multicolumn{1}{l}{} \\ \cmidrule(lr){2-4} \cmidrule(lr){5-7}
\multirow{-2}{*}{$\mathrm{User}_i$} & $\theta_{i1}$                        & $c_{i1}$                       & $\mathrm{ROI}_{i1}$                     & $\theta_{i2}$                        & $c_{i2}$                      & $\mathrm{ROI}_{i2}$                      & \multirow{-2}{*}{Selected Treatment}   \\ \midrule
1                       & \cellcolor[HTML]{C0C0C0}20 & \cellcolor[HTML]{C0C0C0}1 & \cellcolor[HTML]{C0C0C0}20 & 30                         & 2                         & 15                         & 1                    \\
2                       & 15                         & 1                         & 15                         & \cellcolor[HTML]{C0C0C0}36 & \cellcolor[HTML]{C0C0C0}2 & \cellcolor[HTML]{C0C0C0}18 & 2                    \\
3                      & 15                         & 1                         & 15                         & \cellcolor[HTML]{C0C0C0}32 & \cellcolor[HTML]{C0C0C0}2 & \cellcolor[HTML]{C0C0C0}16 & 2                    \\
4                      & \cellcolor[HTML]{C0C0C0}4  & \cellcolor[HTML]{C0C0C0}1 & \cellcolor[HTML]{C0C0C0}4  & 2                          & 2                         & 1                          & 1                    \\ \midrule
5                      & 3                          & 1                         & 3                          & 6                          & 2                         & 3                          &                      \\
6                       & 2                          & 1                         & 2                          & 2                          & 2                         & 1                          &                      \\ \bottomrule
\end{tabular}%
}
\end{table}

\subsubsection{Optimal Treatment Selection Algorithm.}
\label{4.1.2}

Optimal treatment selection algorithm recently proposed by \citet{tu2021personalized} can be used to solve BTS problem without too much modifications. However, this algorithm has two limitations especially for a large-scale BTS problem. 

Firstly, on a large-scale dataset, the author suggested the cohort-level optimization instead of the member-level optimization because of the lack of a large-scale linear programming optimization solver. However, as \citet{tu2021personalized} tested, on low noise level dataset, member-level optimization can generate more personalized estimations. Therefore, we wish to develop a parallel algorithm to solve the member-level optimization for BST problem on a large-scale dataset.

Secondly, in order to realize multi-treatment effect estimation, \citet{tu2021personalized} simply train multiple binary causal forests \cite{wager2018estimation} (MBCF) separately, that is, one causal forest generates the treatment effect estimation for one treatment group versus control group. However, there are two limitations of MBCF:


\begin{itemize}
  \item It is computationally cumbersome to train and maintain many binary causal forests (BCF).
  \item For a user, treatment effect estimation generated by one BCF may belong to different feature space from that of other BCFs, which contradicts the definition of CATE (Eq. \ref{cate}). For instance (Fig. \ref{fig:causalforest}), a user with feature $\{\mbox{age}=30, \mbox{income}=55k\}$ will fall into $\mbox{Leaf}_{11}$ of $\mbox{BCF}_1$ and $\mbox{Leaf}_{31}$ of $\mbox{BCF}_3$, respectively, but $\mbox{Leaf}_{11}$ and $\mbox{Leaf}_{31}$ correspond to different feature values.
\end{itemize}

\subsection{Proposed LBCF Algorithm}
\label{4.3}

We break BTS problem into two steps:
\begin{itemize}
  \item [(i)] Estimate CATE $\theta_{ij}$ at a member-level using just one model.
  \item [(ii)] Get the optimal treatment selection $\mathbf{z^*}$ by solving a constrained optimization problem. 
\end{itemize}

\subsubsection{Unified Discriminative Causal Forest.}
\label{4.2.1}

To overcome the limitations of MBCF (Section \ref{4.1.2}), and to discriminate the treatment effect estimation across multi-treatment, by modifying BCF we design a new multi-treatment causal forest model with the following two properties:
\begin{itemize}
    \item Unified - The model builds only one causal forest, all treatments are split together and follow the same split rules.
    \item Discriminative - The model can discriminate both inter-node heterogeneity and intra-node heterogeneity. 
    
\end{itemize}

\textbf{Splitting Criterion.}
We first propose the split criteria for the new multi-treatment causal forest which consists of \textit{Inter split} and \textit{Intra split}.
\begin{definition}[Inter Split]
\label{rt}
The \textit{Inter split} is a split rule that maximizes the inter-node heterogeneity.
\end{definition}
The \textit{Inter split} rule is to maximize the following (Eq. \ref{eq:UCF}):
\begin{equation}
\label{eq:UCF}
\Tilde{\Delta}_{inter}(\phi_1, \, \phi_2) =  \sum_{l = 1}^2\frac{1}{\left|\{{i : \mathbf{x_i} \in \phi_l}\}\right|}\sum_{j = 1}^K({\sum_{\{i : \mathbf{x_i} \in \phi_l\}} \rho_{ij}})^2.
\end{equation}
Let $\phi_l$ be the feature space associated with a child node $l$. In BCF, $\rho_{i}$ is the pseudo outcome defined in \cite{athey2018generalized}. Intuitively, under treatment $j$, $\rho_{ij}$ can be regarded as $i$'s individual contribution to the total gap between a child node's treatment effect and its parent node's treatment effect. Different from BCF, $\mathbf{\rho_{i}}$ is a vector with length $K$ in our case. The detailed calculation for $\rho_{ij}$ is illustrated in Algorithm \ref{alg:UCF}. Note that for each node's split, $\rho_{ij}$ only needs to be calculated one time, so \textit{Inter split} is quite efficient.
\begin{assumption}
\label{assu:diff_within}
There exist some users whose treatment effects are heterogeneous across multiple treatments.
\end{assumption}

According to assumption \ref{assu:diff_within}, we conclude that if the model can not discriminate the treatment effect across multi-treatment, it is impossible to select the optimal treatment under budget constraint (Section \ref{4.2.2}). Therefore, we propose the following \textit{Intra split} rule:

\begin{definition}[Intra Split]
The \textit{Intra split} is a split rule that maximizes the intra-node heterogeneity, that is to maximize the following:
\end{definition}


\begin{equation}
\label{eq:DSR2}
\Tilde{\Delta}_{intra}(\phi_1, \, \phi_2) = \sum_{l = 1}^2\sum_{j = 1}^K({\hat{\theta}^{(j)}_{\phi_l}-\bar{\theta}_{\phi_l}})^2,
\end{equation}
where $\hat{\theta}^{(j)}_{\phi_l}$ refers to the treatment $j$'s estimated treatment effect and $\bar{\theta}_{\phi_l}$ refers to the average treatment effect of all samples in child node ${\phi_l}$, respectively. Calculation details for $\hat{\theta}^{(j)}_{C_l}$, $\bar{\theta}_{\phi_l}$ are in Algorithm \ref{alg:UCF}. Note, solving for $\hat{\theta}^{(j)}_{\phi_l}$ in each candidate child is quite expensive computationally over all possible axis-aligned splits. 

To sum up the two split rule above and to balance the efficiency and effectiveness especially on a large-scale dataset, we propose a new causal forest, called unified discriminative causal forest (UDCF, Fig. \ref{fig:multiple}), with a two-step split criteria:

\begin{itemize}
    \item [(i)] Let $m$ be a small number relatively to the total number of all possible splits. Pick top $m$ split candidates from all possible splits by \textit{Inter split} rule, according to the result calculated by Eq. \ref{eq:UCF}. 
    \item [(ii)] Select the best split from the $m$ split candidates by \textit{Intra split} rule, according to the result calculated by Eq. \ref{eq:DSR2}.
\end{itemize}    

\textbf{Termination Rule and Treatment Effect Estimation.} Termination rule for tree split is just the same as BCF. After UDCF is constructed, treatment effect estimation $\mathbf{\theta(x_i)}$ can be obtained by fitting $\mathbf{T}$ and  $\mathbf{Y}$ in a weighted least square regression, where the \textit{weight} is the similarity between $\mathbf{x_i}$ and all other samples. We omit the details of this part (see \cite{athey2018generalized}).

\begin{algorithm}[ht!]
\caption{Calculation of $\hat{\theta}^{(j)}_{\phi_l}$, $\bar{\theta}_{\phi_l}$, and $\mathbf{\rho_i}$ for UDCF}
\label{alg:UCF}
\begin{algorithmic}[1]
\STATE Input: outcome $\mathbf{Y}\in \R^{N \times 1}$, treatment $\mathbf{T} \in \R^{N \times K}$. 
\STATE Let $\mathbf{\bar{T}_P} = \frac{1}{N}\sum_{i = 1}^N \mathbf{T_i}$, $\bar{Y}_P = \frac{1}{N}\sum_{i = 1}^N Y_i$, $P$ refers to the parent node.
\FOR {$i = 1$ to $N$}
    \STATE $\mathbf{\hat{T}_i} = \mathbf{T_i} - \mathbf{\bar{T}_P}$, $\hat{Y_i} = {Y_i} - \bar{Y}_P$.
\ENDFOR
\STATE Denote $\mathbf{\hat{T}} = [\mathbf{\hat{T}_i}]_{N \times 1}$, $\mathbf{\hat{T}} \in \R^{N \times K}$ and $\mathbf{\hat{Y}} = [\hat{Y_i}]_{N \times 1}$, $\mathbf{\hat{Y}} \in \R^{N \times 1}$.
\STATE Let $\mathbf{A_P} = \mathbf{\hat{T}^{'}\hat{T}}$,  $\mathbf{\hat{\theta}_P} = \mathbf{A^{-1}_P\hat{T}^{'}\hat{Y}}$.
\STATE Residual $\mathbf{R} = \mathbf{\hat{Y}} - \mathbf{\hat{T}\hat{\theta}_P}$, $\mathbf{R} \in \R^{N \times 1}$ and let $\mathbf{Q} = \mathbf{\hat{T}(A^{-1}_P)^{'}}$, $\mathbf{Q} \in \R^{N \times K}$.
\FOR {$i = 1$ to $N$}
    \STATE $\mathbf{\rho_i} = R_i\mathbf{Q_i}$.
\ENDFOR
\STATE $\hat{\theta}^{(j)}_{\phi_l}$ can be calculated by using the same procedure as ${\mathbf{\hat{\theta}_P}}$.
\STATE $\bar{\theta}_{\phi_l} =  \frac{1}{K}\sum_{j = 1}^K\hat{\theta}^{(j)}_{\phi_l}$
\STATE Output: $\hat{\theta}^{(j)}_{\phi_l}$, $\bar{\theta}_{\phi_l}$, and $\mathbf{\rho_i} \in \R^{1 \times K}$.
\end{algorithmic}
\end{algorithm}

\subsubsection{Budget-constrained Optimization.}
\label{4.2.2}

Based on $\mathbf{\theta}$ learned in Section \ref{4.2.1}, we can formulate our optimization problem given the budget $B$. We wish to get the optimal $\mathbf{z^*}$ by solving the following:

\begin{equation}
\label{eq:ILP}
\begin{aligned}
\max_{\mathbf{z}}~ &\quad \sum_{i=1}^N \sum_{j=1}^K \theta_{ij} z_{ij}  \\
\mbox{s.t.}~~ &\quad \sum_{j=1}^K z_{ij} \leq 1, i=1, ..., N  \\
 &\quad \sum_{i=1}^N \sum_{j=1}^K c_{ij} z_{ij} \leq B  \\
&\quad z_{ij} \in \{0, 1\} , i = 1, ..., N; j=1,..., K,
\end{aligned} 
\end{equation}
where $N$ and $K$ are the number of users and treatments, respectively.
The first constraint $\sum_{j=1}^K z_{ij} \leq 1$ constrains that each user can be assigned at most one treatment. The second constraint $\sum_{i=1}^N \sum_{j=1}^K c_{ij} z_{ij} \leq B$ controls the total cost budget. The last one $z_{ij} \in \{0, 1\}$ indicates that user $i$ can be either assigned treatment $j$ or not. 

Considering hundreds of millions of users on modern Internet platform, we need to solve Eq. \ref{eq:ILP} in parallel. We firstly decompose it into multiple independent sub-problems by utilizing linear relaxation and
dual:

\begin{equation}
\label{eq:lambdaDLP}
\begin{aligned}
\min_{\mathbf\lambda\geq0}~ &\quad L(\lambda) = \sum_{i=1}^N \max(0, \max_{ 1\leq j\leq K} (\theta_{ij} - \lambda c_{ij})) + \lambda B,  \\
\end{aligned} 
\end{equation}
where $\lambda$ is the Lagrange multiplier. It can be proved that $L(\lambda)$ is a convex function with respect to $\lambda$ (see Supplementary Material).

\begin{assumption}
\label{assu:lambda}
$B$ has the following upper bound.
\begin{equation}
B < \sum_{i=1}^N \max_{ 1\leq j\leq K} c_{ij}.
\end{equation}
\end{assumption}


Otherwise, there is a trial solution, i.e., $j^*=\argmax_{j}\theta_{ij}$.

\begin{proposition}
\label{theo:convex}
The optimal value  $\lambda^*$ for $\min\limits_{\lambda \geq 0} L(\lambda)$  must be in the interval $[0, \max\limits_{i, j}\frac{\theta_{ij}}{c_{ij}}]$.
\end{proposition}

\begin{proof}
Consider the derivative of $L(\lambda)$ with respect to $\lambda$. On one hand, for any $\lambda > \max_{i,j}{\theta_{ij}/c_{ij}}$,  $L'(\lambda) = B$, which is greater than $0$. On the other hand,  $L'(\lambda) \to B - \sum_{i=1}^N \max_{ 1\leq j\leq K} c_{ij}$ as $\lambda \to 0$, which is less than $0$ by Assumption \ref{assu:lambda}. Then, the convexity of $L(\lambda)$ implies the global minimum $L(\lambda)$ is obtained at $\lambda^*$ which is between $0$ and  $\max\limits_{i, j}{\theta_{ij}/c_{ij}}$.

\end{proof}



\textbf{Dual Gradient Bisection.} 
To solve Eq. \ref{eq:lambdaDLP}, dual gradient descent is a popular method, but it need to choose an appropriate initial learning rate, which is often computationally intensive especially for large-scale variables. Thanks to Proposition \ref{theo:convex} and convexity of the Lagrange dual, we propose an efficient dual gradient bisection (DGB) method as described in Algorithm \ref{alg:binary}.


\begin{algorithm}[ht!]
\caption{Find Optimal $\lambda^*$ with DGB Method}
\label{alg:binary}
\begin{algorithmic}[1]
\STATE Input: the set of treatment effect values and costs $\{(\theta_{ij}, c_{ij}): i \in \mathcal{U}, j\in \mathcal{J}\}$ for any $i$ user and its any treatment $j$, and budget $B$, and a small constant $\epsilon$. 
\STATE Let $\mathbf{\theta} = \{\theta_{ij}\}_{i \in \mathcal{U}, j\in \mathcal{J}}, \mathbf{c} = \{c_{ij}\}_{i \in \mathcal{U}, j\in \mathcal{J}}$ be the value and cost matrix, compute  $\mathbf{y}(\lambda) := \mathbf{\theta} - \lambda \mathbf{c}$. Denote $L(\lambda) = \sum_{i \in \mathcal{U}} (\max_{j \in \mathcal{J}}\mathbf{y}_{ij}(\lambda)  \vee 0) + B \lambda $ as the dual target.
\STATE Let $\lambda_l = 0, \lambda_r = \max_{i, j}(\mathbf{\theta}/\mathbf{c}), \lambda^* = \frac{\lambda_l + \lambda_r}{2}$.
\STATE Compute the derivative  $L'(\lambda^*)$ at $\lambda = \lambda^*$.

\WHILE{$\lambda_r - \lambda_l  > \epsilon$} 
\IF{$L'(\lambda^*) > 0$}
    \STATE $\lambda_r = \lambda^*$
\ELSE
    \STATE $\lambda_l = \lambda^*$
\ENDIF
    \STATE Let $\lambda^* = \frac{\lambda_l + \lambda_r}{2}$.
    \STATE Compute the derivative  $L'(\lambda^*)$ at $\lambda = \lambda^*$.
\ENDWHILE
\STATE Output: $\lambda^*$
\end{algorithmic}
\end{algorithm}

After obtaining $\lambda^*$ by Algorithm \ref{alg:binary}, for each user $i$, optimal policy $\pi_B$ chooses the treatment $j^*$ according to Eq. \ref{eq:selec}, that is $z_{ij^*} = 1$. Note $j^*=0$ means not to select any treatment at all. 

\begin{equation}
\label{eq:selec}
j^* = 
\begin{cases}
0 &\mbox{if }  \mathop{\max}\limits_{ 1\leq j\leq K} (\theta_{ij} - \lambda^* c_{ij}) \leq 0 \\
\mathop{\arg\max}\limits_{1\leq j\leq K}(\theta_{ij} - \lambda^* c_{ij}) &\mbox{if }  \mathop{\max}\limits_{ 1\leq j\leq K} (\theta_{ij} - \lambda^* c_{ij}) > 0
\end{cases} 
\end{equation}

\textbf{Time Complexity Analysis.} At line 12 of Algorithm \ref{alg:binary}, time complexity of calculating $L'(\lambda^*)$ is $O(N*K)$. At line 5-13, there are at most $\lfloor \log_2{(\frac{\max_{i,j}(\mathbf{\theta}/\mathbf{c})}{\epsilon})}\rfloor + 1$ iterations, which usually is not too large. So the total time complexity of Algorithm \ref{alg:binary} is $O(N*K)$.

\subsubsection{Large-Scale System and Parallelization.}
\label{4.2.3}

For BTS problem, the word ``large-scale'' has two meanings: hundreds of thousands of users and lots of various treatments, both of which make obstacles in applying these techniques to large-scale system. For treatment effect estimation, the proposed UDCF is essentially a random forest which can be run in parallel on modern distributed system. And UDCF is essentially a single model, which can train $K$ treatment groups data together no matter how big $K$ is. For budget-constrained optimization, calculation of gradient $L'(\lambda^*)$ (see Line 12 in Algorithm \ref{alg:binary}) can be decomposed into $\left| \mathcal{U} \right| = N$ independent subproblems, which also can be solved in parallel by the distributed system. In a word, even in a large-scale system, LBCF algorithm can still perform well and run efficiently.

\subsubsection{Overall Algorithm}
\label{4.2.4}  

Based on proposed UDCF and DGB, we sum up our methods in Algorithm \ref{alg:overal}.

\begin{algorithm}[ht!]
\caption{LBCF Algorithm}
\label{alg:overal}
\begin{algorithmic}[1]

\STATE Obtain RCT samples from an online test or a synthetic dataset.
\STATE Generate member-level treatment effect estimation for the outcome under various treatments by UDCF (Section \ref{4.2.1}).
\STATE Solve member-level budget-constrained optimization problem using the proposed DGB (Section \ref{4.2.2}).
\end{algorithmic}
\end{algorithm}








\section{Evaluation}
\label{5}

  

In order to fully conform the performance of our LBCF algorithm, we conduct three kinds of tests: simulation analysis on a public synthetic dataset, offline test on a real-word dataset and online AB test. 

\subsection{Simulation Analysis}
\label{5.1}

\textbf{Synthetic Data.} We use the same method as \citet{tu2021personalized} to generate the synthetic dataset. In this simulation, we assume there are $80$ thousand samples and each sample has three treatments. We maximize the treatment effect of outcome while constraining the cost under the budget $B$. Besides, we also measure the performance of different methods under different levels of noise in data by introducing the uncertainty weight hyperparameter as in \citet{tu2021personalized}.

\textbf{Measurement Metric.} In order to make a convincing result, we also use the same measurement metric as in \citet{tu2021personalized}: normalized mean of individualized treatment effect (ITE) for synthetic data, where
$\mu_{0} = \sum_{i=1}^N \frac{1}{N} Y_i(T_i=0)$ is the average outcome of control group. 



\begin{equation}
\label{ite}
 \tau_{syn} = \frac{\frac{1}{N}\sum_{i = 1}^N\sum_{j = 1}^K(Y_i(T_i=j)-Y_i(T_i=0)){z_{ij}}}{\mu_{0}}.
\end{equation}

\textbf{Simulation Result.} We compare LBCF with the following baseline methods: 
\begin{itemize}
    \item  CTS, ED and Chi of CausalML - Baseline Method
    \item  CT.ST and CF.DT of \citet{tu2021personalized} - Baseline Method
    \item  LBCF - Proposed Method
\end{itemize}

\begin{figure}[http]
    \centering
    \resizebox{0.95\columnwidth}{!}{%
    \includegraphics{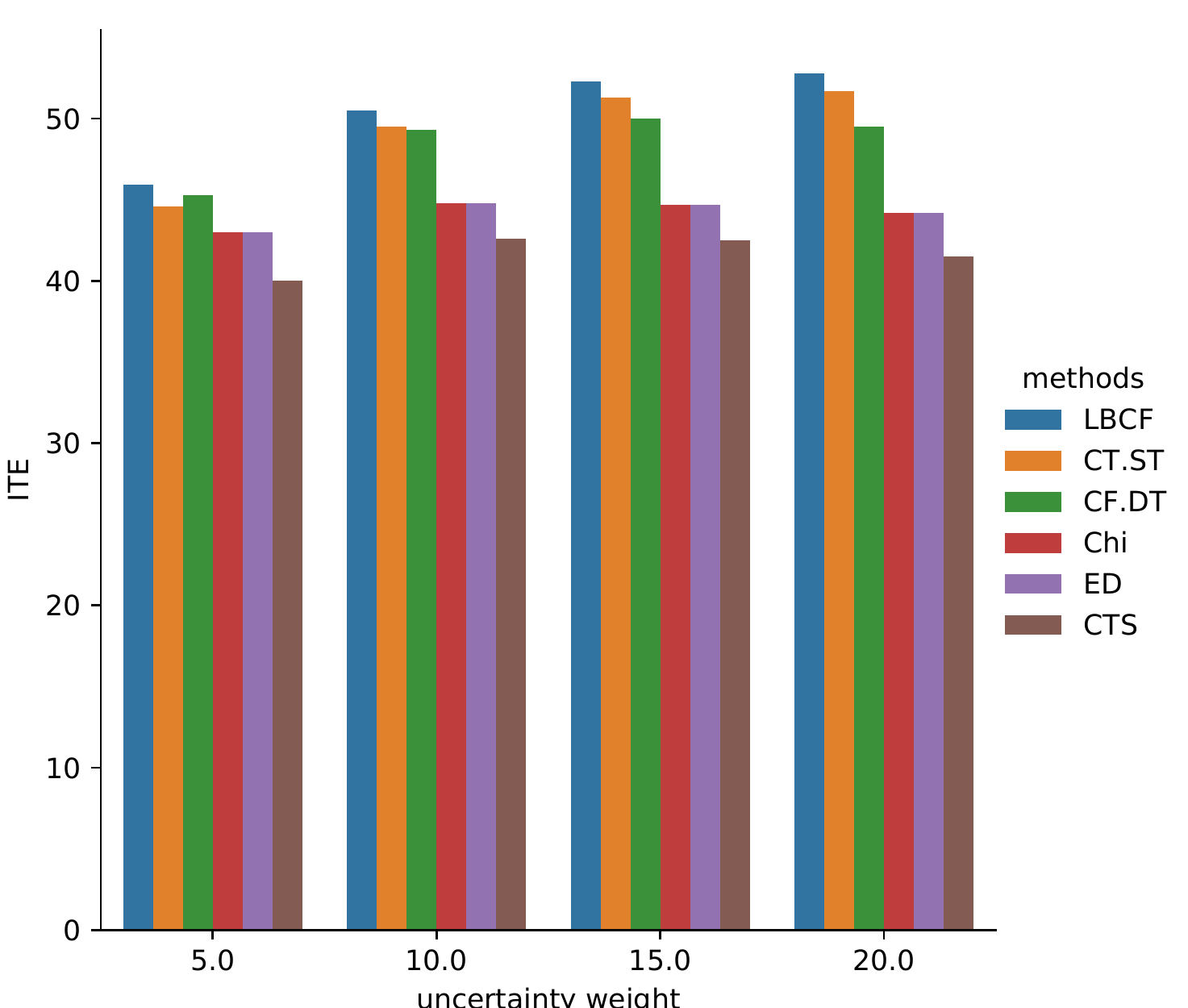}}
    \caption{Simulation results on comparing the proposed algorithm with other baseline methods under different noise levels (uncertainty weights), where x-axis
represents the uncertainty weight, and y-axis represents the mean of individualized treatment effects (ITE) normalized by the average outcome of control group on the synthetic RCT datasets.}
    \label{fig:simresult}
\end{figure}

We report the result in Fig. \ref{fig:simresult}. It's obviously that in any uncertainty weight, LBCF is superior to any other baseline method, which proves the good performance of the proposed algorithm.

\subsection{Offline Test}
\label{5.2}

\textbf{RCT Data Generation.}
On a video app, we allocated $0.1\%$ online traffic to RCT and the RCT ran two weeks to collect enough sample data, which was used as training instances for all the methods compared in Section \ref{5.1}. In the RCT, the campaign operator offered different bonus points to app users of different experiment groups, and offered none to those of control group. Generally speaking, more bonus points encouraged more campaign engagement duration. Finally, the RCT dataset consists of over $100$ thousand app visit instances. Each instance was attributed with associated features (e.g. app visit behaviors, area, historical statistics etc.), outcome (campaign engagement duration) and treatment (bonus points) if applicable (details described in Supplementary Material).

\begin{figure}[http]
    \centering
    \resizebox{0.95\columnwidth}{!}{%
    \includegraphics{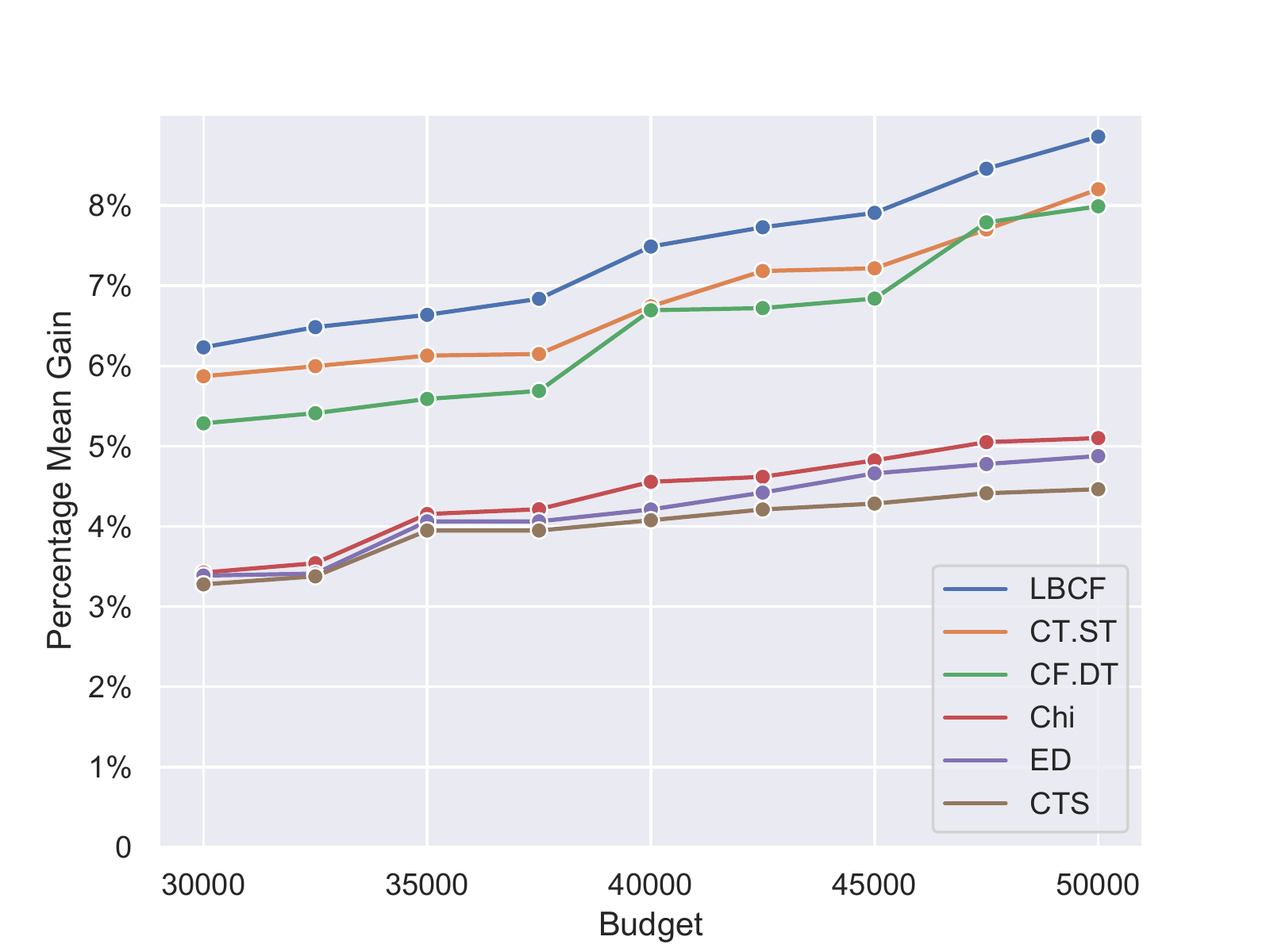}}
    \caption{Offline test results on comparing the proposed algorithm with other baseline methods under different budgets, where x-axis
represents the budgets, and y-axis represents Percentage Mean Gain (PMG) of campaign engagement duration on the real-world RCT datasets.}
    \label{fig:simulation}
\end{figure}

\textbf{Results.}
By utilizing the evaluation metrics PMG described in Eq. \ref{mapg}, we make the same comparison as in Section \ref{5.1} but under different total budget constraints. As in Fig.  \ref{fig:simulation}, the proposed method LBCF achieves largest PMG of user's campaign engagement duration in any budget configuration, which is consistent with the result in Simulation Analysis (see Section \ref{5.1}).

\subsection{Online AB Test}
\textbf{Setups.}
The online AB experiment was conducted for more than $15$ days on a large-scale video app mentioned in Section \ref{5.2}. The goal of the experiment was to evaluate the percentage gain of campaign engagement duration between the experiment groups and a control group (i.e. no incentive group). The experiment groups are all under the same budget constraint $B$ and the control group has no bonus offering. Based on the results of simulation test and offline test above, in this online experiment, we only compare LBCF with the following two methods: CT.ST and CF.DT, due to the high cost of implementing the pipeline and launching online A/B tests. The Linear Programming solver for CF.DT can not handle a large-scale dataset, so we use our proposed DGB instead. Thus in this test, CF.DT is also equivalent to MBCF.DGB (CF is actually the MBCF model).

\begin{itemize}
    \item  Experiment Group: CT.ST of \citet{tu2021personalized} - Baseline Method
    \item  Experiment Group: CF.DT (MBCF.DGB) of \citet{tu2021personalized} - Baseline Method
    \item  Experiment Group: LBCF (UDCF.DGB) - Proposed Method
\end{itemize}

\begin{figure}[t]
    \centering
    \resizebox{0.95\columnwidth}{!}{%
    \includegraphics{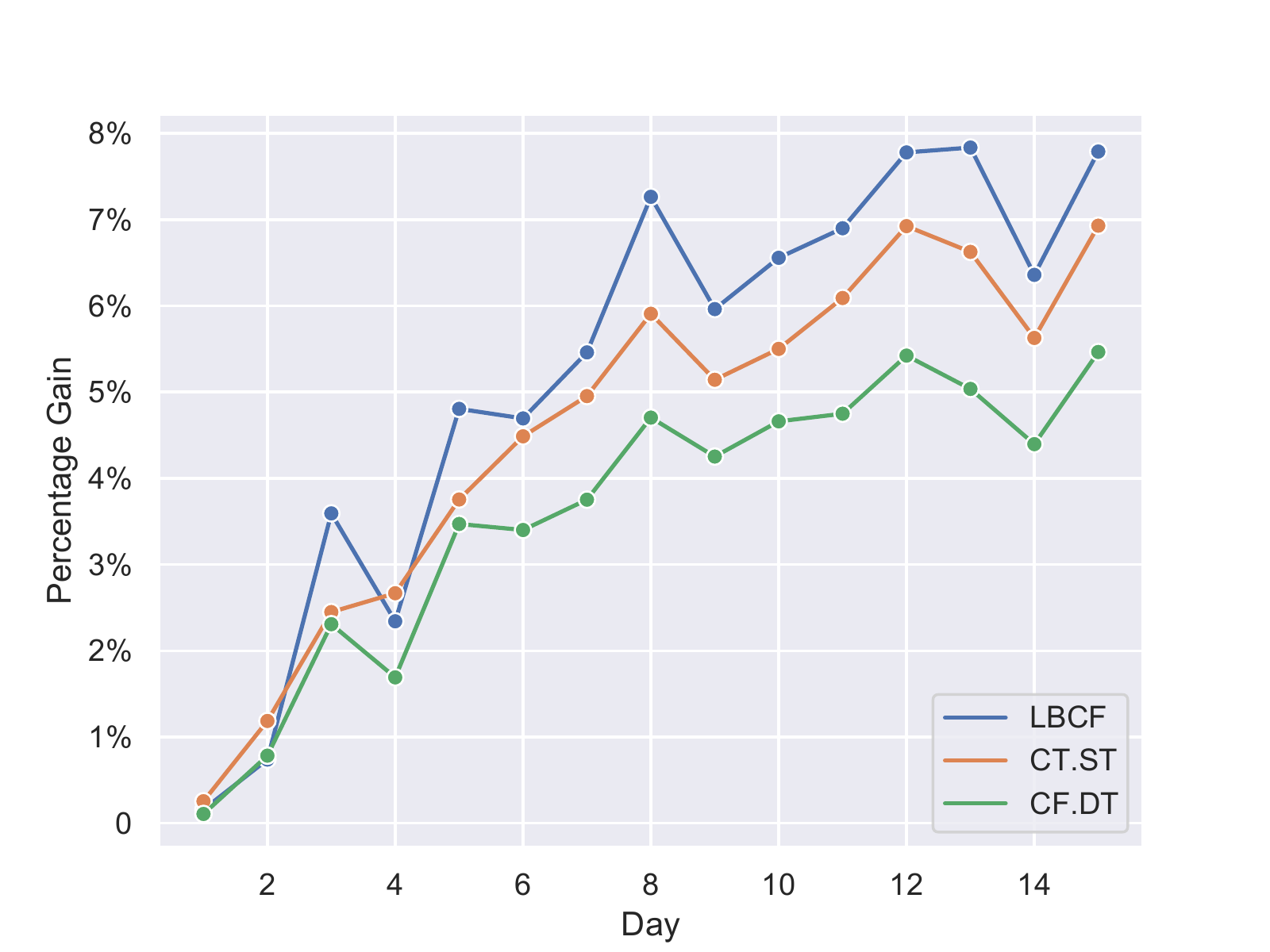}}
    \caption{Online AB test results on comparing the proposed algorithm with other baseline methods during fifteen days, where x-axis
represents the days, and y-axis represents Percentage Gain of campaign engagement duration.}
    \label{fig:online_exp2}
\end{figure}

\textbf{Results.}
Fig.  \ref{fig:online_exp2} presents the results. After the first few days, the proposed LBCF (UDCF.DGB) consistently provides high percentage gain (7.79\% at the last day). One baseline CF.DT (MBCF.DGB) is the worst among groups (5.31\% at the last day), because MBCF is worse than UDCF as we illustrated in Section \ref{method}. Another baseline CT.ST is in the middle (6.87\% at the last day), which is basically consistent with our offline test results in Section \ref{5.2}.

\section{Related Work}
\label{Related Work}

\textbf{Budget-Constrained Treatment Selection.}
For BTS problem, we mainly focus on the tree-based techniques. Existing methods include uplift random forests on Euclidean Distance (ED), Chi-Square (Chi) and Contextual Treatment Selection \cite{zhao2017uplift} (CTS), etc.; Causal tree with stochastic optimization (CT.ST) and Causal forest with deterministic optimization (CF.DT), both of which are the top two methods recommended by \citet{tu2021personalized}. The limitations for these methods are that either they cannot guarantee the same feature space when estimating uplift and often have to train multiple models or they simply adopt a sub-optimal greedy treatment selection policy under a budget limit.
Besides, the budget-constrained optimization MCKP \cite{sinha1979multiple} is a well-known NP-hard problem. In the past few decades, both exact and heuristic algorithms have been studied, including, branch and bound \cite{dyer1998dominance}, tabu search \cite{hiremath2013first}, greedy algorithm \cite{sinha1979multiple} and so on.
Unfortunately, these traditional algorithms were not designed for modern distributed computing. As a result, they cannot solve MCKP on a very large scale dataset (e.g., billions of decision variables). 

\textbf{Evaluation Metric.}
Two kinds of popular evaluation methods exist for BTS problem. Some methods, such as AUUC \cite{gutierrez2017causal, rzepakowski2012decision, soltys2015ensemble}, Qini-curve \cite{radcliffe2007using}, or AUCC \cite{du2019improve} etc., require that similarly scored groups exhibit similar features, which is not necessarily satisfied. Another type of method is the expected outcome metric propose by \cite{zhao2017uplift}. The main limitation of this metric is that the evaluated users are not the whole RCT users, which causes the consumed budget changes with different treatment selection policies. Two policies under different consumed budgets are not comparable.

\section{Discussion}
\label{Conclusion}

SUTVA assumption emphasizes two points: the first point is the independence of each unit. In our case we indeed acknowledge that, SUTVA assumption does not strictly hold on this first point.  For example, two app users may often share the incentive information (i.e. treatment) with each other. Some work recently has been developed to deal with such dependence in data. The second point is that there only exists a single level for each treatment. In our work, we consider the discrete treatments and we guarantee that each treatment is of the same amount of incentives. Therefore, SUTVA assumption can hold on this second point.

Also we would like to add more constraints (e.g. user experience) to the constrained optimization task, and propose an efficient solution to such task even in large-scale system. 

\section{Acknowledgments}
We thank anonymous reviewers for many insightful comments and suggestions.

\bibliographystyle{ACM-Reference-Format}
\balance
\bibliography{main.bib}

\appendix

\section{Supplementary Materials}
\label{sec:reproducibility}

In this section, we describe some further details that can help the reader to reproduce the simulation test in Section \ref{5.1} and offline evaluation in Section \ref{5.2}. We give a proof for the convexity of $L(\lambda)$ in Section \ref{4.2.2}. We also provide the \texttt{Python}, \texttt{C++} and \texttt{R} code that can be used to run all the offline experiments. 

\begin{figure}[http]
    \centering
    \resizebox{0.95\columnwidth}{!}{%
    \includegraphics{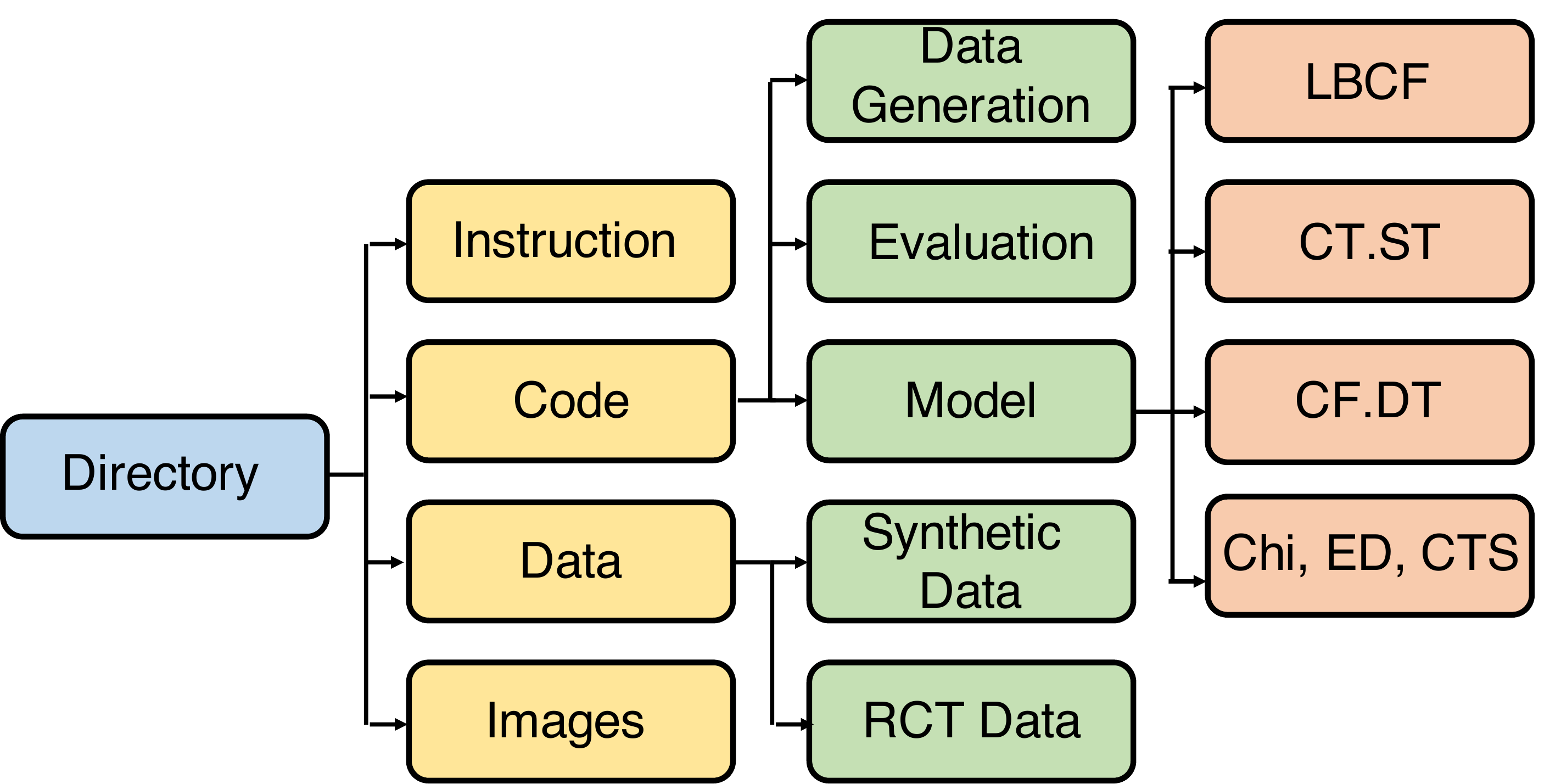}}
    \caption{Directory of Attachment (code and data) }
    \label{fig:online_exp}
\end{figure}

All the data, code files and Image have been uploaded to \href{https://github.com/www2022paper/WWW-2022-PAPER-SUPPLEMENTARY-MATERIALS}{github:} \emph{https://github.com/www2022paper/WWW-2022-PAPER-SUPPLEMENTARY-MATERIALS}, which is organized in the following structure.  (See Fig. \ref{fig:online_exp}).


\subsection{Proof of the convexity of $L(\lambda)$ for Eq. \ref{eq:lambdaDLP}}

\begin{lemma}
$L(\lambda) = \sum_{i=1}^N \max(0, \max_{ 1\leq j\leq K} (\theta_{ij} - \lambda c_{ij})) + \lambda B $ is a convex function of $\lambda$.
\end{lemma}

\begin{proof}
Notice that all linear functions(include constant) are convex and $L(\lambda)$ is obtained by a combination of $\max$ and $+$ operations over a set of linear functions.  Because $\max(\cdot, \cdot)$ and $\cdot + \cdot $ are both operations that keep convexity, the convexity of $L(\lambda)$ can be induced recursively. 
\end{proof}

\subsection{Code}

\textbf{Code Folder} contains three sub-folders: \textbf{Data\ Generation Folder} for synthetic data generation; \textbf{Model Folder} for BTS problem solution methods, containing proposed LBCF method and various baseline methods;  \textbf{Evaluation Folder} for offline evaluation on synthetic data (Section 5.1) as well as on real-world data (Section 5.2).

\textbf{Data Generation Folder} containing codes used for synthetic data generation. We adopted the method mentioned in \citet{tu2021personalized}, and created a self-defined causal Directed Acyclic Graph (DAG). In such DAG, causal relationships are represented by directed arrow, from the cause node pointing to effect node.
Under BTS problem setting, such DAG consists 5 different types of node: heterogeneous variables, denoted as $X$, are generated from underlying causal mechanism with uncertainty effect; unobserved variables, denoted as $U$, are generated from latent distribution; treatment variables, denoted as $T$, are generated from independent sampling; outcome variable, denoted as $Y$, and cost variable, denoted as $C$ are both generated from underlying causal mechanism;

\begin{figure}
	\begin{center}
		\begin{tikzpicture}[>=stealth, node distance=2.5cm]
		\tikzstyle{format} = [draw, thick, circle, minimum size=4.0mm,
		inner sep=1.8pt]
		\tikzstyle{unode} = [draw, thick, circle, minimum size=1.0mm,
		inner sep=0pt,outer sep=0.9pt]
		\tikzstyle{square} = [draw, very thick, rectangle, minimum size=4mm]
		\path[->,  line width=0.9pt]
        node[] (x1) {$X_1$}
        node[ above right of=x1] (x2) {$X_2$}
        node[right of=x2] (x4) {$X_4$}
        node[right of=x4] (c) {$C$}
        node[right of=x1] (x3) {$X_3$}
        node[right of=x3] (y) {$Y$}
        node[right of=y] (u) {$U$}
		node[above right=0.7cm of y] (t) {$T$}	
        (x1) edge[] (x2)
        (x1) edge[] (x3)
        (x2) edge[] (x3)
        (x3) edge[] (x4)
        (x2) edge[] (x4)
        (x2) edge[] (y)
        (x3) edge[] (y)
        (x4) edge[] (y)
        (t) edge[] (y)
        (t) edge (c)
        (x3) edge (c)
        (x4) edge (c)
        (u) edge[dashed] (c)
        (u) edge[dashed] (y);
	    \end{tikzpicture}
	\end{center}
    \caption{Synthetic causal graph with heterogeneity variable (H), unobserved variables (U) treatment variables
(T), outcome variable(Y) and cost variable(C). The edges encapsulate the causal functions including the amount of uncertainty.}
    \label{fig:my_label}
\end{figure}
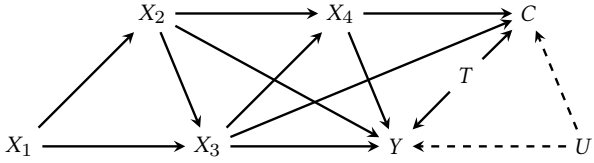

The causal DAG we adopted to generate synthetic data is shown in Fig. \ref{fig:my_label}. We included four heterogeneous variable, one unobserved variables, one treatment variable with four different treatment values, and two outcome variable representing \textit{Value} and \textit{Cost}, respectively. For the purpose of solving BTS problem, \textit{Cost} is set to be a positive random variable.


\textbf{Model Folder} contains implementation of the proposed method LBCF and five other baseline methods, which are used in the simulation test and offline test in Section \ref{5}. For detailed description of baseline methods, we refer the readers to \citet{tu2021personalized} and CausalML documentation.
\begin{enumerate}
\item \textit{LBCF\ Folder} - This folder contains the README file for instruction and code for our proposed method LBCF. Please follow the README file to execute the code.

\item \textit{CT\_ST\ Folder} - This folder contains the README file for instruction and code for CT.ST . Please follow the README file to execute the code.

\item \textit{CF\_DT\ Folder} - This folder contains the README file for instruction and code for CF.DT . Please follow the README file to execute the code.

\item \textit{Chi\_ED\_CTS\ Folder} - This folder contains the README file for instruction and code for Chi, ED and CTS. Please follow the README file to execute the code.

\end{enumerate}

\textbf{Evaluation Folder} contains codes used for synthetic data evaluation and real-world RCT data evaluation. For synthetic data evaluation we adopted ITE measurement proposed in \citet{tu2021personalized} and the result is shown in Section 5.1. For offline test on real-word data, we adopted PMG method described in section 3, corresponding to the result in Section 5.2.

\subsection{Data}

\textbf{Data Folder} contains two sub-folders: \textbf{Synthetic Data Folder} includes training data and testing data generated; \textbf{RCT Data Folder} contains real-world RCT data which has been split into training and testing.

\textbf{Synthetic Data Folder} contains generated simulation data files with two types (training and testing) and four uncertainty levels. These data files are used for offline evaluation mentioned in section 5.1. These data files were generated following the description in section A.2, with different uncertainty weight levels. The unobserved variable was specifically discarded from training data to satisfy the "unobserved" condition.

\begin{enumerate}

\item \textit{Synthetic Data} - Each training sample was attributed with the following:
\begin{itemize}
\item  ID - Column[0], data unique ID.
\item  Heterogeneous Variables - Column[1-4].
\item  Treatment Variable - Column[5], with four different level of treatments.
\item  Outcome Variables - Column[6-7], Value and Cost under BTS problem setting.
\end{itemize}

\end{enumerate}

\textbf{RCT Data Folder} contains web-scale RCT data collected from a video streaming platform which was used as training instances. The dataset records the users' campaign engagement duration (i.e., ``outcome'') in seven randomly enrolled incentive groups, each offered bonuses at different levels (i.e., ``multi-treatment''). In the experiments of Section 5.2, our dataset consists of over 100 K app visit instances. 

Due to the privacy nature of the data, currently we are not able to disclose all the actual data used in our experiments until we get authorization but the practitioner can use the code provided to run this algorithm on other dataset. 
A small part (about 2000 samples) of real-world data are provided in \textbf{Data Folder} just for running the code, \textbf{NOT FOR REPRODUCE THE RESULT} in Section 5.2, and \textbf{ALL THE 2000 SAMPLES HAVE BEEN Encrypted}. As stated in the paper, the complete dataset used in offline test will be released upon authorization.


\begin{enumerate}

\item \textit{RCT} \textit{Data} - In order to let reader learn some meaning of our real-world RCT data features, we only introduce some of them.
\begin{itemize}
    \item  Column[1] - total number of campaign active days in the last 30 days.
    \item  Column[4] - total number of video views in the last 30 days.
    \item  Column[5] - number of fans.
\end{itemize}

\end{enumerate}







\subsection{Utilities}

\begin{enumerate}
\item \textit{DGB.py} - This code provides the implementation of Dual-Gradient-Bisection (DGB) Method proposed in section 4.2.2

\item \textit{Simulation\_analysis.py} - code for synthetic data evaluation .
\item \textit{Simulation\_analysis.ipynb} - detailed step-by-step result for synthetic data evaluation.
\item \textit{Offline\_test.py} - code for real-world RCT data evaluation.
\item \textit{Offline\_test.ipynb} - detailed step-by-step result for real-world RCT data evaluation.


\end{enumerate}


\end{document}